\documentclass[letterpaper, 10 pt, conference]{ieeeconf}  

\IEEEoverridecommandlockouts                              
                                                          
\usepackage{graphics} 
\usepackage{epsfig} 
\usepackage{mathptmx} 
\usepackage{times} 
\usepackage{multirow}
\usepackage{amsmath} 
\usepackage{amssymb}  
\usepackage{algorithmic}
\usepackage{algorithm}
\usepackage{balance}
\usepackage{soul,xcolor}
\usepackage[hidelinks, implicit=false]{hyperref}
\let\labelindent\relax
\usepackage{enumitem}
\usepackage{scalerel}
\setitemize{leftmargin=*}

\usepackage{dsfont}
\usepackage{array}
\usepackage{threeparttable}
\usepackage[noadjust]{cite}
\newcolumntype{P}[1]{>{\centering\arraybackslash}p{#1}}
\newcommand{\ra}[1]{\renewcommand{\arraystretch}{#1}}
\usepackage{siunitx}
\usepackage{hyperref}

\definecolor{ice}{RGB}{140,210,200}

\title{\LARGE \bf
Resilient Legged Local Navigation: Learning to Traverse with Compromised Perception End-to-End
}

\author{Jin Jin$^{\dagger 1}$, Chong Zhang$^{\dagger 1}$, Jonas Frey$^{1,2}$, Nikita Rudin$^{1}$, Matías Mattamala$^{3}$, Cesar Cadena$^{1}$, Marco Hutter$^{1}$
\thanks{$^{\dagger}$ Equal Contribution, listed randomly}
\thanks{$^{1}$ Robotic Systems Lab, ETH Zurich, Zurich, Switzerland}%
\thanks{$^{2}$ Max Planck Institute for Intelligent Systems, Tübingen, Germany.}%
\thanks{$^{3}$ Oxford Robotics Institute, the University of Oxford, UK.}%
\thanks{This work was supported by the Swiss National Science Foundation (SNSF) through project 188596, the National Centre of Competence in Research Robotics (NCCR Robotics), the European Union's Horizon 2020 research and innovation program under grant agreement No 101016970, No 101070405, and No 852044, and an ETH Zurich Research Grant. Jonas Frey is supported by the Max Planck ETH Center for Learning Systems.} 
}

\begin{document}

\maketitle
\thispagestyle{empty}
\pagestyle{empty}

\begin{abstract}

    Autonomous robots must navigate reliably in unknown environments even under compromised exteroceptive perception, or perception failures. Such failures often occur when harsh environments lead to degraded sensing, or when the perception algorithm misinterprets the scene due to limited generalization. In this paper, we model perception failures as invisible obstacles and pits, and train a reinforcement learning (RL) based local navigation policy to guide our legged robot. Unlike previous works relying on heuristics and anomaly detection to update navigational information, we train our navigation policy to reconstruct the environment information in the latent space from corrupted perception and react to perception failures end-to-end. To this end, we incorporate both proprioception and exteroception into our policy inputs, thereby enabling the policy to sense collisions on different body parts and pits, prompting corresponding reactions. We validate our approach in simulation and on the real quadruped robot ANYmal running in real-time ($<$\SI{10}{ms} CPU inference). In a quantitative comparison with existing heuristic-based locally reactive planners, our policy increases the success rate over \SI{30}{\percent} when facing perception failures. Project Page: \url{https://bit.ly/45NBTuh}.
\end{abstract}

\section{Introduction}

Reliable local navigation is essential for autonomous robots in the wild, which typically involves building a real-time local traversability map~\cite{FreyMattamala-RSS-23, artplanner, mattamala2022efficient} or cost map~\cite{wooden2010autonomous,yang2021real} for path planning. Generally, given an accurate map, it is not difficult for existing navigation planners to guide the robot towards the local goal safely. For legged robots that can robustly traverse various terrains~\cite{miki2022learning}, such local planners have become a routine and demonstrated remarkable performance in different tasks~\cite{tranzatto2022cerberus, DavidFanLocalPlanning}.

However, real-world scenarios are complex, and near-perfect perception (\emph{i.e.}, the map representation accurately reflects the environment, with negligible noises or drifts) is not always available. Cameras and LiDARs may not work in dark, rainy, and foggy environments. The vision system may not generalize to the experienced data especially when it is based on neural networks~\cite{kamann2020benchmarking}. Transparent objects and obscured pits are untraversable and difficult to recognize. All these cases lead to invisible obstacles and pits for the navigation module, which we call \textbf{\textit{perception failures}}.

\begin{figure}[t]
\centering
\includegraphics[width=8.5cm]{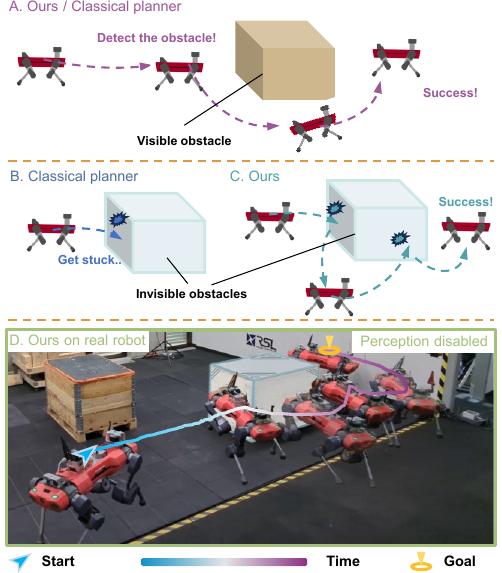}
    \caption{
        An illustration of what happens under perception failures, exemplified by an \textcolor{ice}{invisible obstacle} case. (A) Without perception failures, both classical planners and our navigation policy can function well. (B) When perception failure happens, classical planners get stuck. (C) Our proposed policy can react to perception failures and reach the target. (D) Our proposed policy functions on a real ANYmal-C robot being fully blind to obstacles.
    }
    \label{fig:fig1impress}
    \vspace{-7mm}
\end{figure}

Existing navigation methods struggle with perception failures, as they typically rely solely on exteroceptive sensors to interpret the environment. Most of them construct an explicit map for path planning based on traversability~\cite{FreyMattamala-RSS-23, DavidFanLocalPlanning} or geometry information~\cite{mikielevation2022, Kim-RSS-22} of the surroundings, and some generate twist commands based on raw observations~\cite{hoeller2021learning, truong2023rethinking}. When perception failure happens, these methods can have undesired behaviors, as illustrated in Fig.~\ref{fig:fig1impress}. Few recent works recognize these risks and achieve notable initial success by using anomaly detection and heuristic rules for replanning~\cite{fu2022coupling, ji2022proactive}. However, such manually-designed rules cannot scale well to diverse situations.

To tackle perception failures, we intuitively need certain feedback to recognize the discrepancies between perceived and real environments. In this work, we propose to incorporate locomotion-level observations into navigation, contrasting existing methods that typically decouple navigation from locomotion. These observations include both proprioception and exteroception used by the locomotion module, inspired by Fu~et~al.~\cite{fu2022coupling} using proprioception for collision detection and Ji~et~al.~\cite{ji2022proactive} using exteroception for anomaly detection.


We use end-to-end reinforcement learning (RL) to learn local navigation skills for legged robots. The learned navigation policy generates velocity commands to a pre-existing low-level locomotion policy, and takes low-level observations as part of its inputs. We employ an asymmetric actor-critic design where the actor has corrupted exteroception and memory allowing it to infer the environment, while the critic can access privileged information including the external forces and the perfect map. Compared to heuristic-based local planners~\cite{mattamala2022efficient}, our learned policy exhibits comparable performance under near-perfect perception, and far outperforms the baseline planner under heavily corrupted perception. We also show that legged robots can navigate in the real world even being fully blind.

Our main contributions are:
\begin{enumerate}
    \item The methodology to learn local navigation strategies that can react to and recover from perception failures;
    \item The design of simulation environments and reward functions to train our policy;
    \item Experimental validation of our method both in simulation and real world.
\end{enumerate}

\section{Related Works}

\subsection{Planning-Based Local Navigation}
Classical local planners for legged robots typically use heuristics (for sampling~\cite{wermelinger2016navigation, wellhausen2021rough}, optimization~\cite{mattamala2022efficient}, or rules~\cite{fu2022coupling}) or model-based techniques~\cite{bansal2019-lb-wayptnav, gaertner2021collision} to guide the robot towards a local goal. A representation of the environment must be built from perception, based on geometric information~\cite{mikielevation2022, fankhauser2016universal, frey2022locomotion, wellhausen2021rough, oleynikova2016signed, mainprice2020interior,oleynikova2017voxblox} or semantics~\cite{FreyMattamala-RSS-23, nardi2019actively, wellhausen2019should}. These planners assume near-perfect perception and plan the path with minimal cost based on the provided representation. However, if the perception system fails to identify an obstacle, these planners cannot handle it correctly. In this paper, we assume that perfect perception cannot be guaranteed, and therefore resign to learning a robust local planner that can handle these failure cases while exploiting the provided environment representation when it is reliable.

\subsection{RL-Based Local Navigation}
RL-based local navigation has recently emerged as it is powerful for behavior learning~\cite{prm_rl_icra2018, rudin2022advanced, hoeller2023anymal}, computationally efficient and suitable for mapless navigation~\cite{xie2017towards, hoeller2021learning, Zhao_2021_ICCV, truong2023rethinking, wijmans2023emergence}. However, the lack of a map and the reliance on vision make them vulnerable to local minima and perception failures. For example, Faust~et~al.~\cite{prm_rl_icra2018} and Khaled~et~al.~\cite{nakhleh2023sacplanner} train map-based policies, but assume a near-perfect prebuilt map. Joanne~et~al.~\cite{truong2023indoorsim} uses a context map complimentary to the vision, but cannot handle visual perception failures.

In this paper, we use a traversability map~\cite{mikielevation2022} that can be imperfect, \emph{i.e.}, the map representation can be inconsistent with the environment. We also utilize the proprioceptive information (\emph{e.g.}, joint torques) and the scanned terrain geometry (\emph{i.e.}, the exteroceptive data of the locomotion module) to sense and react to perception failures. 

\subsection{Perception Failures}
Some recent works tackle perception failures in the context of navigation~\cite{fu2022coupling, multitaskfailure,ji2022proactive,antonante2023task, jin2023anomaly}. Antonante~et~al.~\cite{antonante2023task} propose risk estimation of perception failures for autonomous driving. Jin~et~al.~\cite{jin2023anomaly} propose to incorporate anomaly detection into multi-sensor fusion for autonomous driving. Sun~et~al.~\cite{multitaskfailure} and Ji~et~al.~\cite{ji2022proactive} use perception data to predict potential failures. Fu~et~al.~\cite{fu2022coupling} study how proprioceptive data can predict collisions and thereby labeling transparent obstacles in the prebuilt map for navigation planning. In this paper, we explore how legged robots can automatically discover local navigation skills to overcome perception failures in a data-driven manner without heuristic rules. 

Another way around is to improve the perception system. Examples are \cite{FreyMattamala-RSS-23} where the perception module is continuously updated via self-supervision during deployment, and \cite{vasilopoulos2022reactive} where the exteroception can be updated by semantic perceptual feedbacks. However, such methods cannot support real-time reaction to perception failures, as they need to first update their perception module.


\begin{figure}[t]
\centering
\includegraphics[width=7.6cm]{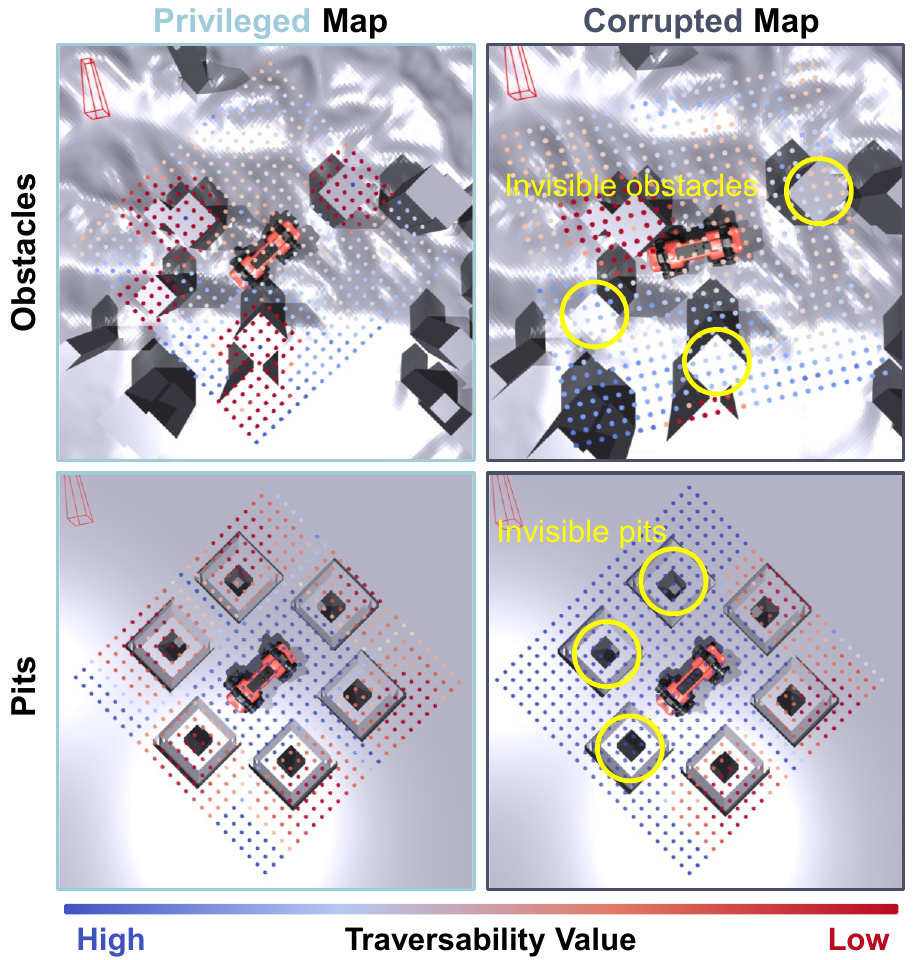}
    \caption{
        An illustration of the privileged map and the corrupted map. The map (visualized by colored dots) provides the terrain traversability around the robot. With corrupted perception, part of the obstacles or pits are invisible to the robot, making their part on the map recognized as traversable.
    }
    \label{fig:travmap}
    \vspace{-6mm}
\end{figure}


\section{Method}
\label{sec:method}

\subsection{Overview}

\begin{figure*}[t]
\centering
\includegraphics[width=176mm, trim={0cm 0cm 0cm 0cm}, clip]{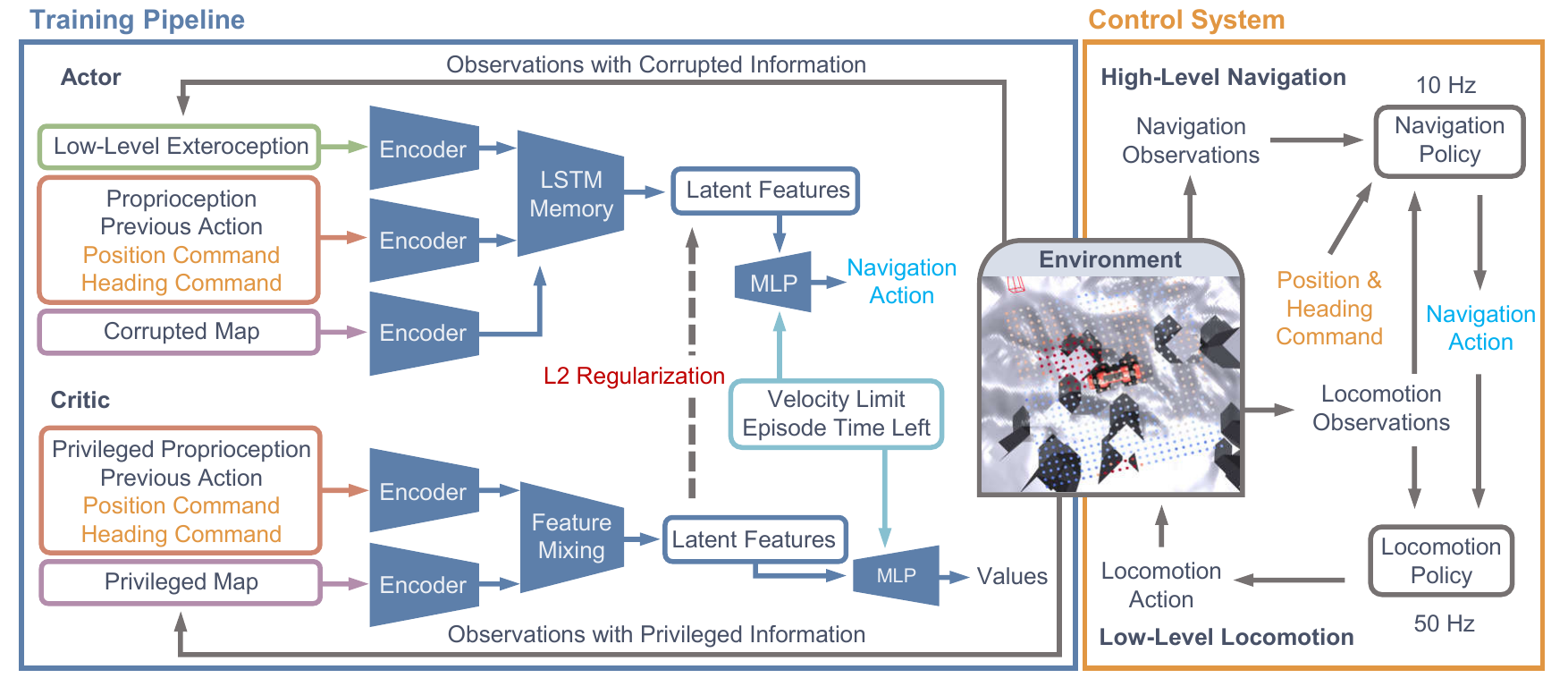}
    \caption{
       An overview of our learning system. \textbf{Left} The actor-critic design of the navigation policy. \textbf{Right:} Our high-level navigation policy generates velocity commands tracked by a low-level locomotion policy. \textbf{Middle:} The simulation environment. The colored points around the robot are the visualized traversability map (as explained in Fig.~\ref{fig:travmap}), which makes the exteroceptive observation of the high-level policy.
    }
    \label{fig:method}
  \vspace{-6.5mm}
\end{figure*}

The objective of our method is to guide the robot to a local target within the given time. To simulate perception failures, we mask part of the obstacles and pits on the terrain to be invisible to the robot's exteroceptive sensing, as shown in Fig.~\ref{fig:travmap}. Besides reaching the target in time, the robot should also reduce base collisions and avoid falls.

An overview of our system is in Fig.~\ref{fig:method}. Given a pre-established low-level locomotion policy~\cite{miki2022learning}, we train a navigation policy that generates velocity commands to be tracked in a hierarchical RL structure. We use the PPO algorithm~\cite{schulman2017proximal} to automatically discover end-to-end navigation strategies capable to cope with perception failures. 

\subsection{Observations}
\label{subsec:obs}

There are four kinds of observations for our navigation policy: the low-level exteroception, the proprioception, the map, and other navigation-related states.

\textbf{Low-Level Exteroception:} The low-level exteroception is the sampled height scan $h_t$ used for the locomotion policy, representing the local terrain perception. Around each foot, there are 52 scanned points, making $h_t$ a 208-dim vector. We refer the readers to~\cite{miki2022learning} for details.

\textbf{Proprioception:} The proprioception consists of the base linear velocity $v_t$, the base angular velocity $\omega_t$, the last five frames of the base acceleration captured at \SI{20}{ms} intervals $\beta_t^{\text{history}}$, the projected gravity on the base $g_t$, the joint positions $q_t$, the joint velocities $\dot{q}_t$, and the last eight frames of joint torques captured at \SI{2.5}{ms} intervals $\tau_t^{\text{history}}$. All of the aforementioned observations are obtained from the state estimator~\cite{bloesch2013state}, and we use the raw IMU data for acceleration measurement. Additional privileged proprioception is provided to the critic consisting of the horizontal external force directions on the base and thighs $F_t^{\text{ext}}$, represented by unit vectors when forces are applied or zero vectors otherwise.

\textbf{Traversability Map:} We use the traversability map of a \SI{3}{m}-wide square area around the robot with \SI{4}{cm} resolution, using the method in~\cite{mikielevation2022}. Such a traversability map is downsampled to a resolution of \SI{12}{cm} as the map observations~$m_t$. The map is illustrated in Fig.~\ref{fig:travmap}. 

\textbf{Navigation-Related States}: Other navigation-related states include the previous action output $a_{t-1}$, the position command $p^c_t$ (\emph{i.e.}, the target position in the base frame), the heading command $\psi^c_t$  (\emph{i.e.}, the target yaw angle in the base frame), the left time of the episode $T-t$ following \cite{rudin2022advanced} where $T$ is the episode length, and the velocity limit $\bar{v}$ allowing users to condition the robot velocity.

\subsection{Asymmetric Actor-Critic with Memory}
\label{subsec:mem}
    We apply asymmetric actor-critic following~\cite{ma2023learning}. The critic network takes privileged proprioception without the joint torques, the privileged map, and other navigation-related states as the inputs. The actor network receives the corrupted map along with the corrupted low-level exteroception, the available proprioception from the state estimator, and other navigation-related states.

    Both networks have a layer that mixes all features from different observations in the latent space. The critic network uses an MLP given that all information is observable from the privileged information. On the other hand, for the actor network, we use an LSTM layer to recurrently infer the partially observable environment following~\cite{wijmans2023emergence}.

\subsection{Implicit State Inference via Regularization}
\label{subsec:reg}
    As the actor has no access to the ground truth perception, we would like to reconstruct the real information from the physical interactions and the corrupted perception. However, our cases do not fit in the popular techniques such as teacher-student~\cite{miki2022learning, lee2020learning, kumar2021rma} and supervised reconstruction~\cite{miki2022learning}. First, in the teacher-student scheme, the privileged teacher can observe and avoid all of the obstacles, which differs from the expected student's behavior of first touching the unseen obstacle and then sidestepping it. Second, the supervised reconstruction scheme involves training decoders to reconstruct the raw privileged information. Yet, in a large map with many unperceived areas, the robot can only reconstruct the very limited part with which it has interacted, making the training of the decoder tricky and hard to generalize.

    In this paper, we propose to use regularization to achieve implicit state inference. We assume that the critic's latent features contain all essential and relevant information for decision-making and that trying to replicate this information can help the actor understand the scene. Therefore, we regularize the LSTM layer's outputs in the actor network towards the latent feature embeddings of the critic network, as is illustrated in Fig.~\ref{fig:method}.

\subsection{Rewards}
\label{subsec:rew}
    There are three types of reward terms involved: the task performance terms, the event penalties, and the regularization terms. All terms are summed up to make the reward function.

    The task performance terms consist of two position tracking terms, one heading tracking term, one standing still term. They are based on the ones in~\cite{rudin2022advanced} and dominate the rewards. Given a short duration $T_r$, these rewards are only received when $t>T-T_r$ regardless of what happens before, thereby avoiding local minima introduced by human priors. They are all in the form of
    \begin{equation}
        r_{\text{task}}=\frac{c_1}{1+\lVert \frac{\text{error}}{c_2} \rVert^2},
        \label{eq:task}\vspace{-1mm}
    \end{equation} 

    where $c_1$ and $c_2$ are positive constants, and the error is the distance-to-target for position tracking, the heading difference for heading tracking, and the velocity scalar for standing still. We have a small $c_2$ for a rigid position tracking term with a large $c_1$ so that the position tracking can be accurate, and a large $c_2$ for a soft position tracking term with a small $c_1$ so that the rewards are less sparse. The heading tracking term and the standing still term are only activated when the distance-to-target is below a threshold besides $t>T-T_r$.

    Regarding event penalties, we have constant negative rewards to penalize the robot when it falls and terminates, or has a collision on the base or thighs. We also constantly penalize the robot for each timestep unless it is close to the target.

    The regularization terms include the ones for energy consumption, action changes and magnitude, velocity tracking errors of the locomotion policy, the overspeeding above the limit, yaw rates, the directional discrepancy between the velocity and the heading, and the velocity commands being too small when the target is distant. 

\subsection{Terrain Curriculum}
\begin{figure}[t]
\centering
\includegraphics[width=85mm, trim={0cm 0cm 0cm 0cm}, clip]{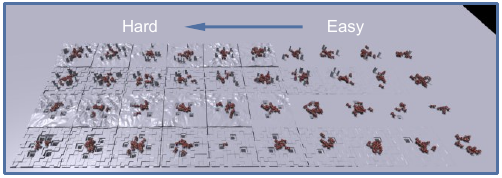}
    \caption{
       We employ a terrain curriculum to facilitate learning. The terrains from top to bottom are obstacles on hilly, obstacles on boxes, pits on hilly, and pits on boxes. The difficulty increases from right to left.
    }
    \label{fig:curriculum}
  \vspace{-6mm}
\end{figure}

\label{subsec:curriculum}
    We employ curriculum learning to tackle challenging scenarios following~\cite{rudin2022learning} and~\cite{rudin2022advanced}. We have discrete boxes in~\cite{rudin2022learning} and realistic hilly terrains in~\cite{Chong2022terrain_benchmark} as the basic terrains to avoid overfitting the flat terrain, and add obstacles and pits on them, as shown in Fig.~\ref{fig:curriculum}. With increasing difficulty, the roughness and step height of the terrain increases as well as the number of obstacles and pits. A robot gets promoted to higher difficulty levels if it reaches the target position within the episode, and gets demoted otherwise.

\begin{figure}[t]
\centering
\includegraphics[width=7.1cm, trim={0cm 0.9in 0cm 1in}, clip]{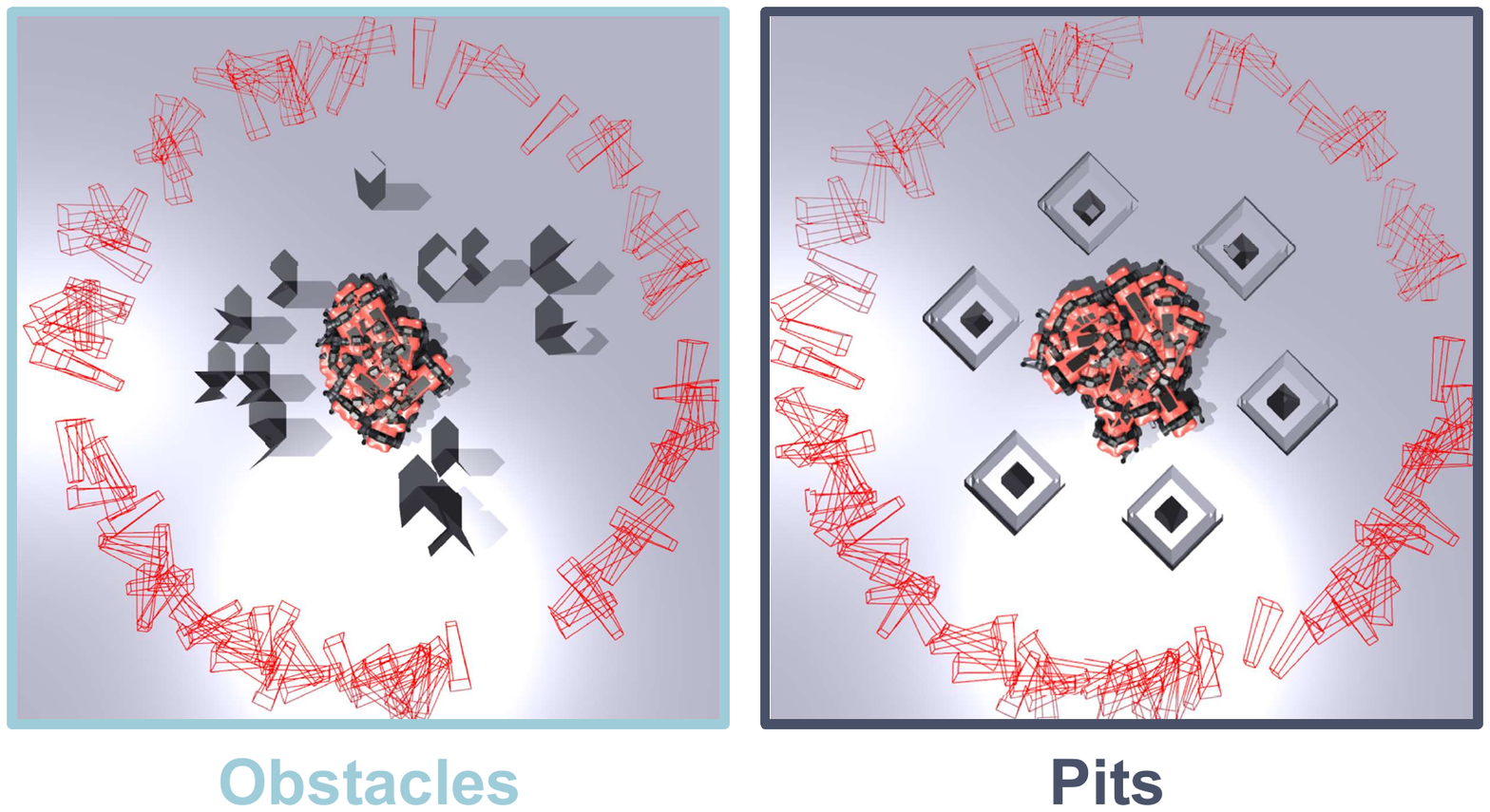}
    \caption{Test terrains in simulation. The red boxes are the targets. 100 robots are tested together.
    }
    \label{fig:evalenv}
  \vspace{-7mm}
\end{figure}

\section{Experimental Setup}

\subsection{Environments}
We verify our methodology on the quadruped ANYmal robot both in simulation and in the real world. We use the Isaac Gym~\cite{Isaacgym} for simulation. To simulate perception failure scenarios, we control the ratio of visible obstacles and pits (also called "visibility") during training and evaluation, where \SI{100}{\percent} indicates that all obstacles or pits are correctly reflected within the traversability map and the height scan.

\subsection{Training Settings}
For training, we consider the following settings:
\begin{enumerate}[leftmargin=*]
    \item \textbf{Ours:} As is presented in Sec.~\ref{sec:method}, we use all the observations we list in the methodology, and also all the techniques including the memory and the implicit state inference via regularization. \SI{50}{\percent} of the obstacles and pits are visible \textit{during training}.
    \item \textbf{Oracle:} This setting is the same as \textbf{Ours} except that \SI{100}{\percent} of the obstacles and pits are visible \textit{during training}. 
    \item \textbf{No Proprioception (NoPro):} This setting differs from \textbf{Ours} in that we set the base acceleration, the projected gravity, and the joint states (positions, velocities, and torques) in proprioception to zero. These observations are typically not used in traditional navigation modules.
    \item \textbf{No Low-Level Exteroception (NoExt):} This setting differs from \textbf{Ours} in that we mask the low-level exteroception with zeros.
    \item \textbf{No LSTM Memory (NoMem):} This setting differs from \textbf{Ours} in that we replace the LSTM layer in the actor network with an MLP layer. 
    \item \textbf{No Regularization (NoReg):} This setting differs from \textbf{Ours} in that we cancel the regularization proposed in Sec.~\ref{subsec:reg} for implicit state inference.
\end{enumerate}
Each setting is run 3 times with different random seeds for statistics, which supports the significance of our performance by P Values. Besides these trained policies, we also use a heuristic-based local planner in \cite{mattamala2022efficient}, a representative for classicial local planners, as a baseline (abbreviated as \textbf{Planner}).

\subsection{Evaluation and Metrics}
\label{subsec:metric}

In simulation, we build two test terrains (one for obstacles and one for pits) for quantitative evaluation of policies, as shown in Fig.~\ref{fig:evalenv}. The visibility of the obstacles and pits is selected from $\{\SI{100}{\percent},\SI{50}{\percent},\SI{0}{\percent}\}$, making 6 simulated test environments: Obstacles-\SI{100}{\percent}, Obstacles-\SI{50}{\percent}, Obstacles-\SI{0}{\percent}, Pits-\SI{100}{\percent}, Pits-\SI{50}{\percent}, Pits-\SI{0}{\percent}.

We use the following metrics for quantitative comparison:
\begin{enumerate}[leftmargin=*]
    \item \textbf{Success Rate.} We count the proportion of robots that can reach the target within \SI{18}{s} which is the episode length.
    \item \textbf{Average Base Collisions.} We count the base collisions per \SI{0.1}{s} for each robot and sum these values over time. This metric is only applicable to obstacles.
    \item \textbf{Average Time Cost.} We count the average time cost for each robot to reach the target. For failed episodes, the time cost is calculated as \SI{30}{s} (the qualitative results hold for any value $\ge$\SI{18}{s}).
\end{enumerate}


\subsection{Randomized Collision Bodies for Obstacles}
\label{subsec:randbody}

To trigger different collision cases \textit{during training} and enhance our policy's generalization ability, we randomize the collison body of the base on obstacle terrains. By doing so, we can learn policies that can react to collisions on different parts of the body in different directions, as demonstrated in Sec.~\ref{subsec:realworld} and video attachment.

\section{Results and Analyses}

\subsection{Emergent Behaviors in Simulation}

\begin{figure}[t]
\centering
\includegraphics[width=8.5cm, trim={0cm 0.2cm 0cm 0.cm}, clip]{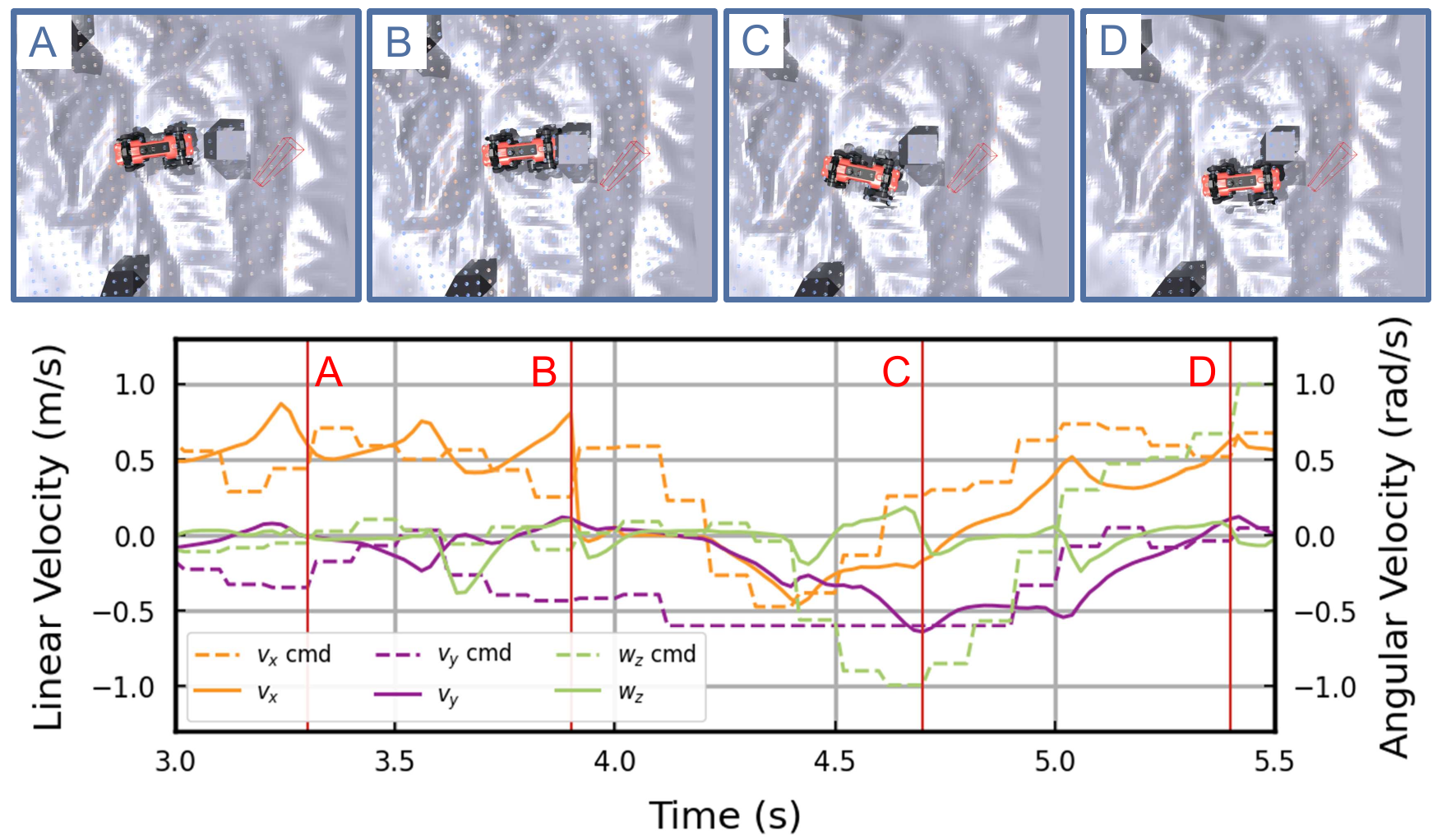}
    \caption{Emergent behaviors of obstacle avoidance after a collision in simulation. The policy can react to the collision and make a sidestep.
    }
    \label{fig:data_plot_obs}
  \vspace{-3mm}
\end{figure}

\begin{figure}[t]
\centering
\includegraphics[width=8.5cm, trim={0cm 0cm 0cm 0.cm}, clip]{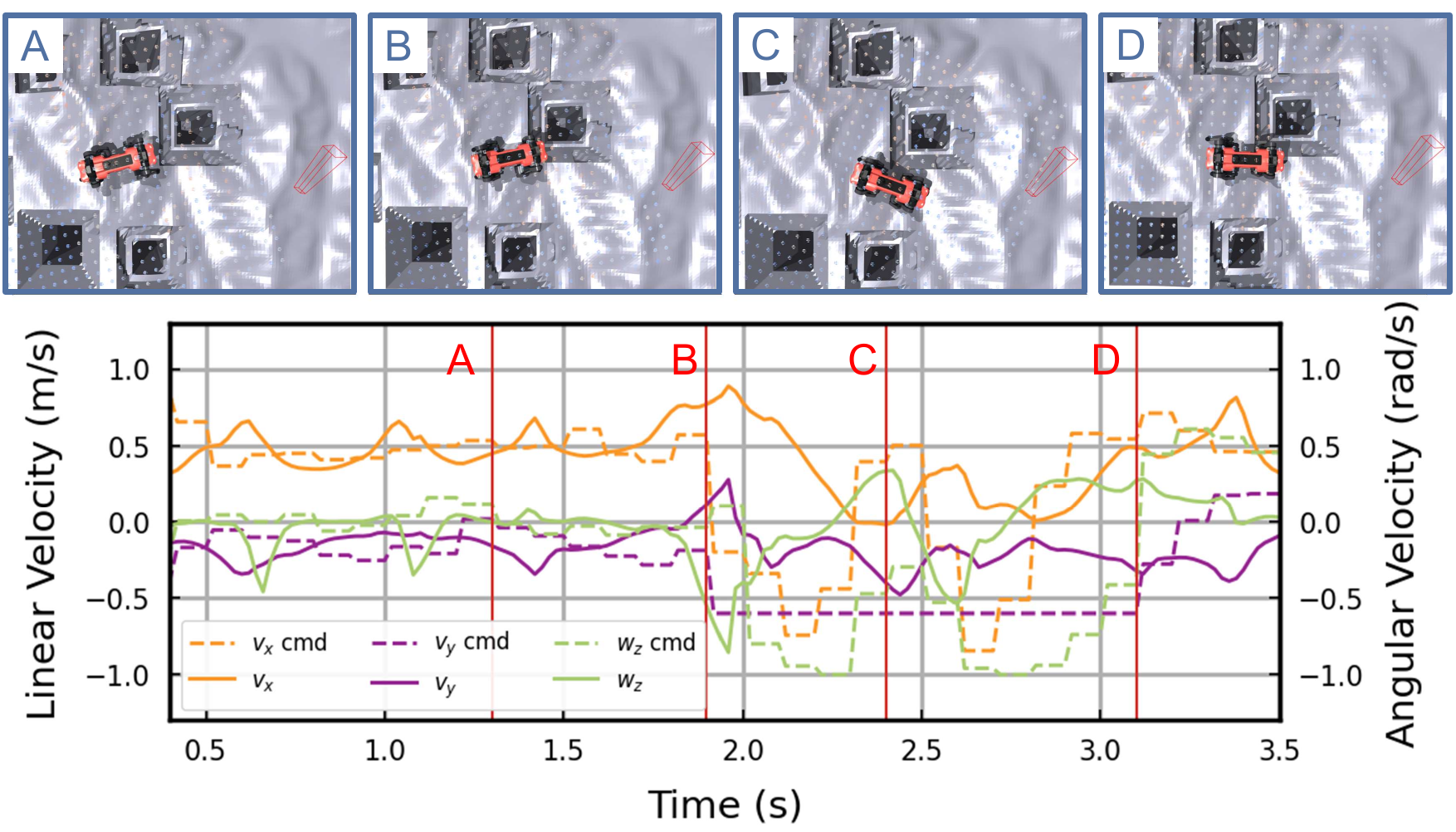}
    \caption{Emergent behaviors of recovery from an invisible pit in simulation. The policy can react to the missed step and drag the robot out of the pit.
    }
    \label{fig:data_plot_pits}
  \vspace{-6mm}
\end{figure}

Our learned policy demonstrates reactive behaviors against perception failures on rough terrains in simulation, as shown in Fig.~\ref{fig:data_plot_obs} and Fig.~\ref{fig:data_plot_pits}. When colliding with an unseen obstacle, our policy gives a backward longitudinal velocity and a large lateral velocity to sidestep the obstacle. When the robot missteps into an invisible pit, the policy generates velocity commands directing away from the pit until the robot regains stability.

\subsection{Comparison Results}

\begin{table*}[tb]
\ra{1.2}
\centering
\caption{Comparison Results in Simulation}
\label{tab:comparison}
\resizebox{177mm}{!}{%
\begin{tabular}{ccccclcccclcccc}
\hline
\multirow{2}{*}{Environment} & \multicolumn{4}{c}{Success Rate (\%)} &  & \multicolumn{4}{c}{Avg. Base Collisions} &  & \multicolumn{4}{c}{Avg. Time Cost (s)} \\ \cline{2-5} \cline{7-10} \cline{12-15} 
 & Oracle & Ours & P Value & Planner &  & Oracle & Ours & P Value & Planner &  & Oracle & Ours & P Value & Planner \\ \cline{1-5} \cline{7-10} \cline{12-15} 
 
Obstacles-\SI{100}{\percent} &  93.3$\pm$1.5  & 93.7$\pm$5.5  &  0.538 & \textbf{100} &  & 1.4$\pm$2.1 & 0$\pm$0 & 0.156 & 0.2 &  & 9.5$\pm$0.4 & 9.9$\pm$0.8 & 0.750 & \textbf{6.9} \\

Obstacles-\SI{50}{\percent} &  54.0$\pm$6.9  & \textbf{ 92.0$\pm$1.7 } & \textless{}0.001 & 61 &  & 32.1$\pm$4.3 & \textbf{1.9$\pm$0.9} & \textless{}0.001 & 23.8 &  & 18.1$\pm$1.3 & \textbf{10.5$\pm$0.8} & \textless{}0.001 & 15.7\\

Obstacles-\SI{0}{\percent} &33.3$\pm$6.4 & \textbf{84.0$\pm$1.7} & \textless{}0.001 & 53 &  & 48.3 $\pm$ 13.6 & \textbf{4.6$\pm$1.2} & 0.003 & 24.9 &  & 22.5$\pm$1.5 & \textbf{12.3$\pm$0.7} & \textless{}0.001 & 18 \\ \cline{1-5} \cline{7-10} \cline{12-15} 

Pits-\SI{100}{\percent} & \textbf{99.7$\pm$0.6} & 98.0$\pm$1.0 & 0.967 & \textbf{100} &  & \multicolumn{1}{c}{\multirow{3}{*}{0}} & \multicolumn{1}{c}{\multirow{3}{*}{0}} & \multicolumn{1}{c}{\multirow{3}{*}{None}} & \multicolumn{1}{c}{\multirow{3}{*}{0}} &  & 7.4$\pm$0.4 & 8.7$\pm$0.3 & 0.995 & \textbf{6.1} \\

Pits-\SI{50}{\percent} & 74.0$\pm$4.4 & \textbf{92.7$\pm$2.3} & 0.001 & 77 &  & \multicolumn{2}{c}{} & \multicolumn{2}{c}{} &  & 13.0$\pm$1.3 & \textbf{9.8$\pm$0.9} & 0.013 & 11.5 \\

Pits-\SI{0}{\percent} & 57.3$\pm$3.1 & \textbf{93.0$\pm$5.3} & \textless{}0.001 & 57 &  & \multicolumn{2}{c}{} & \multicolumn{2}{c}{} &  & 18.7$\pm$0.6 & \textbf{9.8$\pm$1.5} & \textless{}0.001 & 16.4 \\ \hline
\end{tabular}%
}
\vspace{-3mm}
\end{table*}

We compare the proposed \textbf{Ours} with the baselines \textbf{Oracle} and \textbf{Planner} in simulation. The results are presented in Table~\ref{tab:comparison}, with the mean, std, and P values of the metrics. A P value smaller than 0.05 indicates that \textbf{Ours} is significantly better than \textbf{Oracle} regarding the metric in the environment.

According to the results, all of the policies perform well when the visibility is \SI{100}{\percent}, and the \textbf{Planner} achieves a perfect \SI{100}{\percent} success rate. However, as the visibility decreases, \emph{i.e.}, when perception failures increase, \textbf{Ours} drop performance much slower than the other two, and significantly outperform them. \textbf{Ours} can have $>\SI{80}{\percent}$ success rates even when the perception is fully corrupted, and generalize well to \SI{0}{\percent} and \SI{100}{\percent} visibility despite being trained with \SI{50}{\percent} visibility. These results indicate that the navigation policy cannot learn to react to perception failures without being exposed to them, and the locomotion policy cannot overcome perception failures by its own. 

It is also worth mentioning that, the \textbf{Planner} receives a \SI{8}{m}$~\times~$\SI{8}{m} map, but our policies only receives a \SI{3}{m}~$\times$~\SI{3}{m} one, which makes our policies more vulnerable to local minima and fail in rare cases even with perfect perception. However, expanding the map observation is non-trivial as it increases the dimensionality and can make the training data-inefficient or even failed.

\subsection{Ablation Studies}

\begin{figure}[t]
\centering
\includegraphics[width=8.5cm, trim={0cm 0cm 0cm 0cm}, clip]{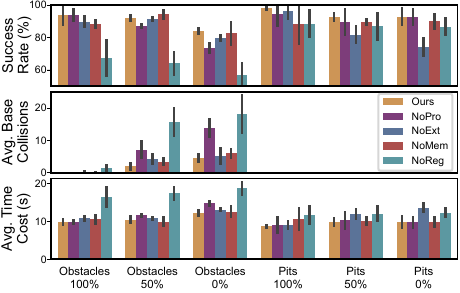}
   \caption{Results of ablation studies. \textbf{Ours} outperforms others in the metrics.}
   \label{fig:ablation}
 \vspace{-6mm}
\end{figure}

We evaluate different settings for ablation studies in simulation. The results are presented in Fig.~\ref{fig:ablation}, with error bars describing the std values. We draw the following conclusions based on the ablation results:
\begin{enumerate}[leftmargin=*]
    \item The proprioception as part of the observations is generally beneficial to the robustness against perception failures, as shown by the \textbf{NoPro} results.
    \item The locomotion-level exteroception can help improve the performance on pits, but demonstrates fewer effects for obstacles, as shown by \textbf{NoExt} results. We hypothesize that other observations are sufficient to detect the collisions, but insufficient to detect a missed step in time, while the exteroception may show some discrepancy with other observations.
    \item The memory only slightly contributes to robustness, as shown by \textbf{NoMem} results. This does not comply with the conclusions in~\cite{wijmans2023emergence} that memory is the key to blind navigation. This can be explained by that memory is not necessary when collisions can be detected by other observations, \emph{e.g.}, the bug algorithm~\cite{ng2007performance} without memory. 
    \item The regularization for implicit state inference is \textit{critical} to the performance in every scene, as shown by the \textbf{NoReg} results. Without such regularization to guide the learning, the policy may converge to a suboptimal solution.
\end{enumerate}
Briefly speaking, both the locomotion-level observations and all techniques we propose contribute to the navigational robustness against perception failures.

\subsection{Real-World Results}
\label{subsec:realworld}

\begin{figure}[t]
\centering
\includegraphics[width=8.5cm, trim={0cm 0cm 0cm 0cm}, clip]{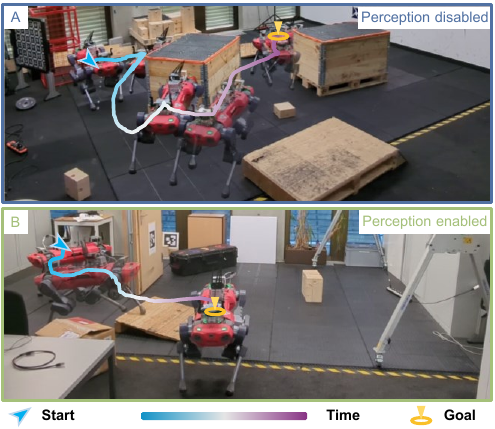}
    \caption{
       (A) Our proposed policy can reach the target when the robot is blind. It can react to collisions on the thighs, the base front, the legs, and the base side. (B) Our proposed policy can traverse rough terrains when perception is enabled.
    }
    \label{fig:realobs}\vspace{-4mm}
\end{figure}

\begin{figure}[t]
\centering
\includegraphics[width=7.8cm, trim={0cm 0cm 0cm 0cm}, clip]{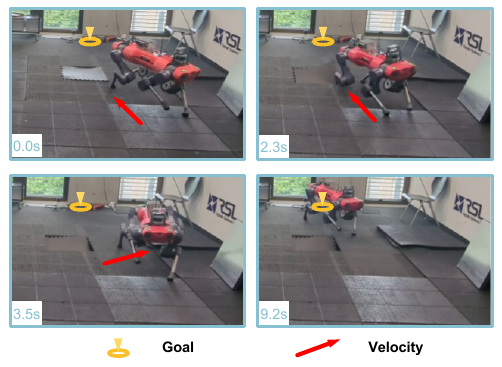}
    \caption{Snapshots of our policy recovering from an invisible pit. After one missed step in the pit, the policy reacts with velocity commands to the right, and guides the robot slowly towards the target after recovery. Red arrows indicate the moving direction of the robot.
    }
    \label{fig:realpit}
  \vspace{-6mm}
\end{figure}

In the real world, we compare \textbf{Ours} against \textbf{Planner} in our lab. \textbf{Our} policy can perform well around obstacles not only when there is near-perfect perception, but also when being fully blind, as shown in Fig.~\ref{fig:realobs}. Under perception failures, the robot first collides with the obstacle and then tries to sidestep around it. Such reactions can happen with collisions on different parts of the robot and in different directions, including the base, the thighs, and the legs (including feet). In comparison, the \textbf{Planner} always gets stuck in front of invisible obstacles (motions in our video attachment).

\textbf{Our} policy can also recover from an invisible pit, as shown in Fig.~\ref{fig:realpit}. It can react at the moment when the robot's foot goes deep under the perceived terrain and tries to drag the robot out of the pit with lateral velocity commands. After the recovery, the policy can keep guiding the robot toward the target. In comparison, the \textbf{Planner} cannot react to such cases and the locomotion policy cannot recover by itself.

\section{Limitations and Future Works}

Despite our policy's generalization to different collision geometries, we find it cannot handle out-of-distribution mapping noises. Real world mapping corruption is hard to model. They are dependent on the SLAM module as well as on the various sensor modalities used to measure the environment.

Besides, our policy always avoids untraversable regions on the map. However, the perception module can generate artifacts, \emph{i.e.}, non-existent obstacles or pits which can stop the robot from approaching the target. Hence, it is of great interest if we can train a policy to actively explore these areas and explicitly revise the map allowing it to take into account false positive and false negative perception failures. Therefore, a closer coupling of the locomotion and navigation policy may be desirable to achieve certain probing patterns to safely explore the traversability of the environment. 

Further research is needed to identify the optimal map representation provided to the local planner, which is currently limited to a 2D map, as well as improving the performance under perfect perception. Further, we will investigate scaling the proposed approach to fully rough terrains in the wild. 



\bibliographystyle{IEEEtran}
\balance
\bibliography{IEEEabrv,references}

\end{document}